\documentclass[conference]{IEEEtran}

\usepackage[nocompress]{cite}
\usepackage{amsmath,amssymb,amsfonts}
\usepackage{graphicx}
\usepackage{textcomp}
\usepackage{xcolor}

\def\BibTeX{{\rm B\kern-.05em{\sc i\kern-.025em b}\kern-.08em
    T\kern-.1667em\lower.7ex\hbox{E}\kern-.125emX}}
\usepackage{booktabs}

\usepackage{algpseudocode}

\algtext*{EndIf}
\algtext*{EndFor}
\algtext*{EndProcedure}

\usepackage{eso-pic}
\AddToShipoutPictureBG*{%
  \AtPageLowerLeft{%
    \put(\LenToUnit{0.5\paperwidth},\LenToUnit{0.32in}){%
      \makebox[0pt][c]{\parbox{0.86\paperwidth}{\centering\scriptsize
        \copyright~2026 IEEE. Personal use of this material is permitted.
        Permission from IEEE must be obtained for all other uses, in any
        current or future media, including reprinting/republishing this
        material for advertising or promotional purposes, creating new
        collective works, for resale or redistribution to servers or lists,
        or reuse of any copyrighted component of this work in other works.
        Accepted at the 38th IEEE International Conference on Tools with
        Artificial Intelligence (ICTAI 2026).}}}}}

\begin{document}

\title{ARMED: Adaptive Retrieval-Augmented Medication Recommendation\\
}

\author{\IEEEauthorblockN{1\textsuperscript{st} Kyungkeon Chung}
\IEEEauthorblockA{\textit{UX Factory, Inc.} \\
Seongnam-si, South Korea\\
kunnyjung@uxf.ai}
\and
\IEEEauthorblockN{2\textsuperscript{nd} Se Un Park}
\IEEEauthorblockA{\textit{UX Factory, Inc.} \\
Seongnam-si, South Korea\\
s4park123@uxf.ai}
\and
\IEEEauthorblockN{3\textsuperscript{rd} Hyejin Lee}
\IEEEauthorblockA{\textit{UX Factory, Inc.} \\
Seongnam-si, South Korea\\
hyejin@uxf.ai}
\and
\IEEEauthorblockN{4\textsuperscript{th} Junyoung Park}
\IEEEauthorblockA{\textit{UX Factory, Inc.} \\
Seongnam-si, South Korea\\
junyoung@uxf.ai}
}

\maketitle

\begin{abstract}
Medication recommendation requires jointly considering diagnoses, comorbidities, longitudinal patient history, and treatment interactions. Although recent retrieval-augmented generation approaches using language models have shown promise for clinical recommendation, most existing systems rely on fixed retrieval pipelines that cannot adapt to diverse clinical retrieval scenarios and may introduce noisy or irrelevant evidence. We propose \textbf{ARMED}, an adaptive clinical RAG framework that dynamically expands retrieval according to retrieval sufficiency and current-admission relevance. ARMED first observes the initial retrieval outcome and selectively performs supplementary or fallback multi-query retrieval when additional evidence is needed. The framework also incorporates a relevance gate that activates only clinically relevant same-patient history and organizes retrieved evidence into structured query-level clinical evidence. Experiments on MIMIC-IV demonstrate that ARMED consistently outperforms previous static retrieval baselines and improves overall medication recommendation quality. By combining locally runnable small language models with adaptive retrieval, ARMED supports medication recommendation without relying on external language model APIs or task-specific fine-tuning.
\end{abstract}


\section{Introduction}
Medication recommendation is an important task in electronic health record (EHR)-based clinical decision support~\cite{gamenet, mesin}. Since multiple medication classes are often prescribed simultaneously during hospital admissions, medication recommendation is inherently a complex multi-label prediction problem~\cite{cognet}. Effective recommendation requires consideration of clinical information such as diagnoses, comorbidities, longitudinal patient history, prior treatments, and treatment interactions~\cite{acdnet}. Previous studies have therefore proposed various deep learning-based methods using structured EHR information~\cite{leap,gamenet,safedrug,mesin,cognet}.

Recently, large language models (LLMs) have been explored for clinical reasoning and medical recommendation tasks~\cite{pmcllama, medpalm}. However, medication recommendation requires clinically grounded and label-constrained outputs, making LLM-only approaches vulnerable to unsupported medication label generation and hallucinated recommendations~\cite{medicalhallucination}. Retrieval-augmented generation (RAG)-based clinical recommendation frameworks, such as MEDPLAN~\cite{medplan}, PACE-RAG~\cite{pacerag}, and TreatRAG~\cite{treatrag}, address this limitation by grounding generated outputs in retrieved clinical evidence. Such retrieval-grounded systems are also relevant to hospital environments where privacy requirements, internal-network deployment, and restrictions on external API use limit reliance on cloud-based proprietary models~\cite{sok, dennstadt2025implementing}.

Despite these advances, many clinical RAG methods still rely on static or pre-defined retrieval pipelines. Static retrieval may introduce practical failure modes such as evidence dilution, majority bias, and history noise, where loosely related retrieved cases or irrelevant previous visits negatively affect the final recommendation. Although PACE-RAG progressively refines evidence through a multi-stage pipeline, its retrieval flow remains largely pre-defined rather than dynamically controlled according to patient-specific retrieval sufficiency or history relevance~\cite{pacerag}.

In this work, we propose \textbf{ARMED}, an adaptive clinical RAG framework for medication recommendation. ARMED dynamically adjusts retrieval behavior according to patient-specific clinical context, selectively utilizes clinically relevant same-patient history, and generates recommendations based on dynamically retrieved evidence. Inspired by recent adaptive RAG studies~\cite{selfrag,crag,adaptiverag,flare,rqrag,searchr1}, our framework incorporates retrieval sufficiency and patient history relevance to mitigate noisy evidence injection and unsupported recommendations.

The main contributions of this work are as follows:
\begin{itemize}
    \item We propose \textbf{ARMED}, an adaptive clinical RAG framework that dynamically adjusts retrieval strategies according to patient-specific clinical context.
    \item We introduce a retrieval sufficiency-based controller that determines whether to expand retrieval based on the initial retrieval outcomes.
    \item We mitigate history noise by selectively utilizing clinically relevant same-patient history based on current admission relevance.
    \item We design an adaptive retrieval-based medication recommendation framework with consideration for compact and locally deployable language model environments.
\end{itemize}

The remainder of this paper presents related work in Section~\ref{sec:related_work}, the proposed framework in Section~\ref{sec:method}, the experimental setup in Section~\ref{sec:experimental_setup}, the results in Section~\ref{sec:results}, and the conclusion in Section~\ref{sec:conclusion}.

\section{Related Work}
\label{sec:related_work}

Medication recommendation has been widely studied as an EHR-based multi-label prediction task~\cite{gamenet,cognet}. Conventional deep learning approaches learn relationships among diagnoses, procedures, medications, and longitudinal patient histories to predict subsequent prescriptions~\cite{gamenet,safedrug}. Although these methods have improved prescription prediction, they typically require task-specific training and do not explicitly retrieve patient-level clinical evidence during inference. This limits their flexibility when medication label spaces, patient populations, or clinical workflows change. 

Beyond task-specific learning from EHR data, prior clinical
cases can provide evidence for new decisions, an approach
that predates modern RAG systems.
Case-based reasoning has been studied in clinical decision
support as a way to reuse previous cases~\cite{bichindaritz2006case},
while structure-mapping theory explains analogical reasoning
through relational correspondences between
situations~\cite{gentner1983structure}.
Clinical RAG follows a related case-reuse perspective when
retrieving patient records to support recommendations.
TreatRAG~\cite{treatrag} retrieves similar historical admissions,
MEDPLAN~\cite{medplan} incorporates patient evidence into clinical
planning, and PACE-RAG~\cite{pacerag} refines evidence and
prescription decisions through a multi-stage pipeline.

Although PACE-RAG progressively refines evidence through
a pre-defined pipeline~\cite{pacerag}, ARMED determines
whether retrieval expansion is needed based on the initial
retrieval outcome and selectively incorporates same-patient
history according to current-admission relevance.

Recent general-domain RAG studies have increasingly treated retrieval as a dynamically controllable process rather than a fixed preprocessing step. FLARE~\cite{flare}, Self-RAG~\cite{selfrag}, CRAG~\cite{crag}, Adaptive-RAG~\cite{adaptiverag}, and RQ-RAG~\cite{rqrag} have explored retrieval necessity estimation, retrieval quality assessment, and query- or context-dependent retrieval control. In this work, we apply this adaptive retrieval perspective to medication recommendation.

\begin{figure}[t]
\centering

\begin{minipage}{\columnwidth}
\hrule
\vspace{0.4em}

\algrenewcommand{\algorithmiccomment}[1]{%
  \hfill{\footnotesize$\triangleright$~#1}}
  
\begin{algorithmic}[1]

\Require admission $x$, visits $V_x$, retrieval DB $\mathcal{D}$, similarity thresholds $T,T_{\mathrm{low}}$, minimum hits $K$
\Require $R(\mathcal{D},q)$: retrieve candidate $c$ and score $s$ from $\mathcal{D}$, maximum rewrites $R_{\max}$

\State $H \gets \Call{HistoryGate}{x,V_x}$
\State $E \gets \Call{AdaptiveRetrieve}{x,\mathcal{D}}$
\State $E_q \gets \Call{BuildQueryEvidence}{E,H}$
\State \Return $\Call{LLMRecommendation}{x,E_q}$

\Statex

\Procedure{AdaptiveRetrieve}{$x,\mathcal{D}$}

    \State $(C_0,S_0) \gets \{(c,s) \in R(\mathcal{D},x):s \geq T\}$

    \If{$|C_0| \ge K$}

        \If{top and mean of $S_0$ are high}
            \State \Return $C_0$ \Comment{Single-query sufficient}
        \Else
            \State \Return $C_0 \cup \Call{MultiQuery}{x,\mathcal{D}}$ \Comment{Supplement}
        \EndIf

    \Else
        \State \Return $\Call{MultiQuery}{x,\mathcal{D}}$ \Comment{Fallback}
    \EndIf

\EndProcedure

\Statex

\Procedure{MultiQuery}{$x,\mathcal{D}$}

    \State $\{q_1^{0},\dots,q_M^{0}\} \gets \Call{SLMDecompose}{x}$
    \State $E \gets \emptyset$

    \For{each decomposed query $q_i^{0}$}

        \For{$r \gets 0$ to $R_{\max}$}

            \State $C \gets \{c : (c,s) \in R(\mathcal{D},q_i^{r}),\, s \geq T\}$

            \If{$|C| \ge K$}
                \State $E \gets E \cup C$
                \State \textbf{break}

            \ElsIf{$r < R_{\max}$} \Comment{Rewrite}
                \State $q_i^{r+1} \gets \Call{SLMRewrite}{q_i^{r}}$

            \Else \Comment{Rescue}
                \State $E \gets E \cup \{c : (c,s) \in R(\mathcal{D},q_i^{0}),\, s>T_{\mathrm{low}}\}$ 
            \EndIf

        \EndFor

    \EndFor

    \State \Return $E$

\EndProcedure

\end{algorithmic}

\vspace{0.4em}
\hrule

\end{minipage}

\caption{Adaptive retrieval workflow.}
\label{fig:adaptive_retrieval_algorithm}
\end{figure}

\begin{figure*}[t]
  \includegraphics[width=\textwidth]{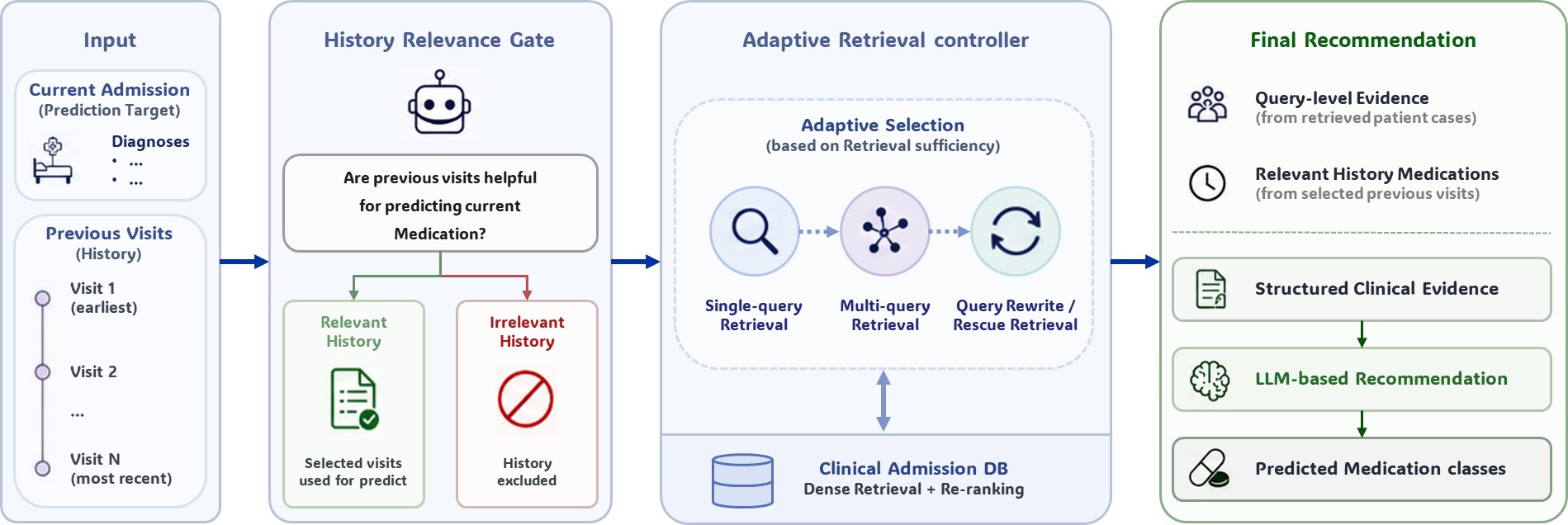}
  \caption{Overview of ARMED.}
  \label{fig:framework}
\end{figure*}

\section{Method}
\label{sec:method}

ARMED combines same-patient history from previous visits
with cross-patient evidence from similar admissions in the
retrieval database. Given the current admission diagnoses,
a history relevance gate selects relevant previous visits
(Section~\ref{sec:method_history}), while adaptive retrieval
starts with a single query and expands when the retrieved
evidence is insufficient (Section~\ref{sec:method_retrieval}).
Both evidence sources are structured to guide the final
LLM-based medication recommendation
(Section~\ref{sec:method_recommendation}).
Figs.~\ref{fig:adaptive_retrieval_algorithm}
and~\ref{fig:framework} illustrate the adaptive retrieval
workflow and overall framework, respectively.

\subsection{Relevant Patient History Selection}
\label{sec:method_history}

Same-patient history can provide useful signals for medication
recommendation, particularly for chronic diseases or recurrent
admissions. However, prior visits may contain diagnoses and
prescriptions unrelated to the current admission, introducing
irrelevant context when included indiscriminately.

ARMED therefore employs an SLM-based history relevance gate
that jointly considers the current admission diagnoses and
the diagnoses and medications from each prior visit to perform
binary relevance classification. Only relevant visits are
retained to reduce irrelevant historical context in the final
LLM prompt. Their medications are incorporated as separate
history evidence alongside cross-patient retrieval evidence
for the final recommendation.

\begin{figure*}[t]
  \includegraphics[width=\textwidth]{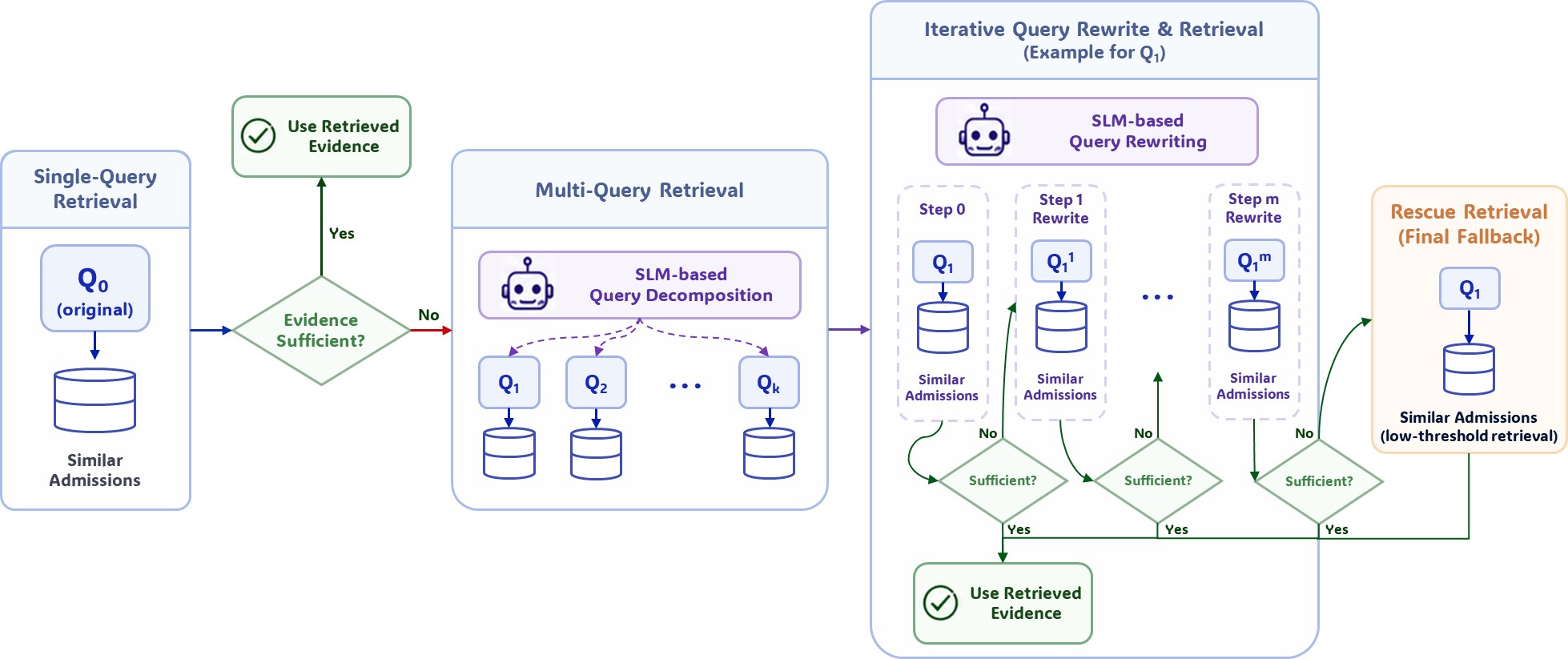}
  \caption{Adaptive retrieval with single-query selection, multi-query expansion, rewriting, and rescue.}
  \label{fig:adaptive_framework}
\end{figure*}

\subsection{Adaptive Retrieval Policy}
\label{sec:method_retrieval}

Retrieval sufficiency cannot be reliably predicted from
query complexity alone, as it also depends on the cases
available in the retrieval database. Complex admissions
may have strongly matched cases, whereas seemingly simple
admissions may retrieve insufficient evidence. ARMED therefore
performs an initial search and uses the observed results
to determine whether retrieval expansion is needed.

As illustrated in Fig.~\ref{fig:adaptive_framework}, ARMED
first retrieves using the current admission's full diagnosis
list as a single query. Diagnosis texts are encoded as dense
sentence embeddings and compared using cosine similarity.
Cases with similarity scores of at least the threshold are accepted.
Retrieval sufficiency is assessed using the number of accepted
hits and their maximum and mean similarity scores.

High-confidence results are used without expansion
(``single-query sufficient''). If the minimum hit requirement is met but the high-confidence
criteria are not, ARMED retains the initial results and adds
multi-query evidence (``supplement'').
Otherwise, multi-query evidence replaces the initial results
(``fallback''). For both routes, an SLM decomposes the diagnosis
list into focused clinical queries, each retrieved independently.

Sub-queries with insufficient evidence are rewritten by an SLM
and retried up to the maximum rewrite count (``rewrite'').
If evidence remains insufficient, a rescue pass uses the
original sub-query with a lower similarity threshold
(``rescue''). Sub-queries with no results meeting the applicable
threshold contribute no evidence. This policy expands retrieval only when the initial
evidence is insufficient.

\begin{figure}[t]
\centering
\fbox{
\parbox{0.94\linewidth}{
\raggedright

[System / task description ...]

\smallskip

Given the target patient's current diagnoses
\textbf{\{CURRENT\_DIAGNOSES\}},
query-level retrieval evidence
\textbf{\{QUERY\_LEVEL\_EVIDENCE\}},
and optional same-patient history
\textbf{\{HISTORY\_EVIDENCE\}},
recommend medication class labels for the target admission.

\smallskip

Select labels only from the provided candidate set; do not introduce unsupported labels or alter label names. ...

}
}
\caption{Abstracted final recommendation prompt. Ellipses indicate prompt content omitted for brevity.}
\label{fig:prompt_example}
\end{figure}

\subsection{Final Recommendation}
\label{sec:method_recommendation}

Retrieved cases are not directly inserted into the LLM prompt as raw retrieval outputs. Instead, ARMED organizes retrieved diagnosis-prescription evidence into a structured query-level evidence format for the final recommendation stage.

For each retrieval query, the framework summarizes retrieved patient cases together with their associated diagnoses and prescription patterns. Repeatedly retrieved medications are additionally aggregated at the query level. This query-specific organization helps preserve the clinical context associated with each retrieval query rather than relying solely on simple frequency-based medication aggregation.

When relevant same-patient history is activated, selected history medications are incorporated as separate history evidence. The final LLM recommendation is then generated using the current diagnosis information together with structured query-level retrieval evidence and selected history evidence (Fig.~\ref{fig:prompt_example}).

\section{Experimental Setup}
\label{sec:experimental_setup}

\subsection{Dataset and Evaluation Splits}
In this study, we constructed an admission-level medication recommendation dataset using MIMIC-IV v3.1~\cite{mimiciv,physionet}. MIMIC-IV contains real-world clinical records deidentified by the dataset providers before release, including removal of patient identifiers and patient-specific date shifting that preserves within-patient temporal intervals. Access was obtained through PhysioNet~\cite{physionet} after completing the required training and credentialing process and agreeing to the data use agreement. We selected MIMIC-IV because its linked diagnosis, prescription, and longitudinal admission records support evaluation of both cross-patient retrieval and same-patient history utilization.

Diagnosis information was represented using ICD diagnosis titles, while prescription information was converted into ATC level 3 medication class names through RxNorm~\cite{rxnorm} and RxClass~\cite{rxclass} mapping based on the April 2026 NIH RxNorm Prescribable release. The final prediction target was defined as the medication class label set for each admission, formulating the task as a multi-label recommendation problem in which multiple medication classes may exist within a single admission.

After removing admissions without diagnosis information or mapped medication labels, the final dataset consisted of 462,084 admissions from 196,194 unique patients. The dataset contained 390 unique medication labels in total, with an average of 12.8 diagnoses and 12.4 medication labels per admission.

The retrieval corpus comprised 461,084 admissions from
195,818 patients with 390 unique medication labels, while
the inference set comprised 1,000 admissions from 376 patients
with 248 labels. The two sets were patient-disjoint by
\texttt{subject\_id}, with no patient or admission overlap.
All inference labels were present in the retrieval corpus;
thus, the evaluation focused on recommending medication
labels supported by existing prescription evidence rather
than generating unseen labels.

We used two evaluation settings to separate overall performance evaluation from history selection analysis. In the admission-level
setting, we evaluated 1,000 admissions from 376 patients.
These comprised 376 first admissions without prior history
and 624 subsequent admissions with prior same-patient history.

The patient-latest setting was used only for the history
selection analysis (Section~\ref{sec:history_selection}),
evaluating the latest admissions of the 167 patients with
prior history and excluding the 209 patients without history.
All other MIMIC-IV evaluations used the admission-level setting.
In both settings, only admissions preceding the evaluated
admission were available as history candidates.

\subsection{Implementation Details}
For final recommendation generation in ARMED, we used Qwen3-8B~\cite{qwen3}. In addition, lightweight control tasks such as diagnosis decomposition, query rewriting, and history relevance gating were performed using Qwen2.5-3B~\cite{qwen25}. All LLM/SLM inference was conducted using Ollama~\cite{ollama}.

For dense retrieval embedding, we used the all-MiniLM-L6-v2 sentence-transformer model~\cite{sentencebert,minilm} and constructed a FAISS~\cite{faiss} vector database using the full retrieval corpus. 
In addition, a MiniLM-based cross-encoder reranker trained on MS MARCO~\cite{msmarco} was used to improve retrieved case ranking quality. 

The initial single-query retrieval accepted cases with similarity scores of at least 0.80. With at least two accepted cases, retrieval terminated without expansion if both the maximum and mean scores reached 0.88 and 0.84, respectively; otherwise, supplementary multi-query retrieval was performed. Fewer than two accepted cases triggered multi-query fallback. We allowed up to three alternative rewrites per decomposed query. These thresholds were fixed experimental settings, without a claim of optimality.

We evaluated final medication recommendations as a multi-label prediction task using admission-averaged precision, recall, and F1. These metrics assess agreement between predicted and recorded medication-label sets, rather than the relevance of retrieved documents.

For admission $i$, let $Y_i$ and $\hat{Y}_i$ denote the ground-truth and predicted medication-label sets, respectively. Labels were matched exactly after lowercasing and normalizing punctuation and whitespace. We computed
\begin{equation}
P_i = \frac{|Y_i \cap \hat{Y}_i|}{|\hat{Y}_i|}, \quad
R_i = \frac{|Y_i \cap \hat{Y}_i|}{|Y_i|}, \quad
F1_i = \frac{2|Y_i \cap \hat{Y}_i|}{|Y_i|+|\hat{Y}_i|}.
\end{equation}

Ratios with zero denominators were assigned zero. Each metric was averaged over admissions, not medication classes. Reported F1 is the mean of per-admission F1 scores, not the harmonic mean of averaged precision and recall.

All experiments were conducted on a workstation equipped with an AMD Ryzen 7 9800X3D CPU, an NVIDIA GeForce RTX 5080 GPU with 16 GB of VRAM, and 64 GB of DDR5 memory, running 64-bit Windows 10.

\section{Results}
\label{sec:results}

\subsection{Overall Performance}

\begin{table}[t]
\caption{Overall medication recommendation performance.}
\label{tab:overall_performance}
\centering
\setlength{\tabcolsep}{3.8pt}
\begin{tabular}{lcccc}
\toprule
Method & F1 & 95\% CI & Precision & Recall \\
\midrule
History
& 0.2926
& [0.255, 0.326]
& 0.2859
& 0.3352 \\

Static RAG
& 0.4093
& [0.397, 0.422]
& 0.3841
& 0.5084 \\

Static RAG + History
& 0.4472
& [0.432, 0.463]
& 0.4121
& 0.5669 \\

TreatRAG
& 0.4009
& [0.378, 0.425]
& 0.4002
& 0.4712 \\

PACE-RAG
& 0.3557
& [0.312, 0.397]
& 0.4157
& 0.3553 \\

ARMED
& \textbf{0.5103}
& \textbf{[0.491, 0.531]}
& \textbf{0.4910}
& \textbf{0.6263} \\
\bottomrule
\end{tabular}
\end{table}

Table~\ref{tab:overall_performance} reports overall medication recommendation performance using the same Qwen3-8B backbone and 1,000 inference admissions from 376 patients for all methods.
For F1, we report 95\% patient-cluster bootstrap confidence intervals computed from 10,000 bootstrap resamples.

The History baseline uses the current admission's diagnoses and available same-patient history without external retrieval. Static RAG uses the complete diagnosis list of the current admission as a single query for fixed dense retrieval and incorporates the retrieved cases into the LLM context, without same-patient history. Static RAG + History combines this retrieved evidence with same-patient history. Both history-using baselines include available history without relevance filtering. Static RAG achieved higher F1 than History, and adding same-patient history further improved F1.

TreatRAG~\cite{treatrag} was reimplemented following its
original lexical n-gram Jaccard retrieval pipeline and
achieved slightly higher precision but slightly lower F1
than Static RAG. We evaluated PACE-RAG~\cite{pacerag}
using the authors' public implementation. Consistent with
the history-dependent performance reported in that study,
PACE-RAG achieved F1 scores of 0.5269 with prior
same-patient history and 0.0716 without history in our
evaluation, compared with 0.5462 and 0.4508 for ARMED,
respectively. Overall, ARMED achieved the highest F1,
precision, and recall among all evaluated methods.

\begin{table}[t]
\caption{Statistical significance of F1 improvements.}
\label{tab:significance}
\centering
\setlength{\tabcolsep}{4.8pt}
\begin{tabular}{lccc}
\toprule
Comparison
& $\Delta$F1
& 95\% CI
& $p_{\mathrm{Holm}}$ \\
\midrule
vs.\ Static RAG + History
& +0.0631
& [0.0522, 0.0735]
& $<0.001$ \\

vs.\ TreatRAG
& +0.1094
& [0.0916, 0.1279]
& $<0.001$ \\

vs.\ PACE-RAG
& +0.1546
& [0.1271, 0.1855]
& $<0.001$ \\
\bottomrule
\end{tabular}
\end{table}

Paired patient-cluster bootstrap tests using 10,000 resamples
showed that ARMED achieved significantly higher F1 than
Static RAG + History, TreatRAG, and PACE-RAG
(Table~\ref{tab:significance}), with Holm-adjusted
$p<0.001$ across all three comparisons.

\begin{table}[t]
\caption{Mean medication prediction counts per admission.}
\label{tab:prediction_counts}
\centering
\begin{tabular}{lrrrr}
\toprule
Method & Predicted & TP & FP & FN \\
\midrule
Static RAG + History & 15.834 & 6.490 & 9.344 & 6.136 \\
ARMED & 14.652 & \textbf{7.146} & \textbf{7.506} & \textbf{5.480} \\
\bottomrule
\end{tabular}
\end{table}

Table~\ref{tab:prediction_counts} shows that ARMED predicted
fewer medication labels than Static RAG + History while
increasing true positives (TP) and reducing false positives
(FP) and false negatives (FN), indicating more accurate
label selection rather than a larger prediction set.

\subsection{Relevant History Selection Analysis}
\label{sec:history_selection}

\begin{table}[t]
\caption{Effect of same-patient history selection.}
\label{tab:history_selection}
\centering
\begin{tabular}{lccc}
\toprule
Method & F1 & Precision & Recall \\
\midrule
w/o History & 0.4387 & 0.4279 & 0.5293 \\
Always-History & 0.4783 & 0.4629 & 0.5757 \\
ARMED & \textbf{0.4792} & \textbf{0.4635} & \textbf{0.5777} \\
\bottomrule
\end{tabular}
\end{table}

Table~\ref{tab:history_selection} evaluates how same-patient
history selection affects final medication recommendation
on the latest admissions of the 167 test patients with prior
history. The \textit{w/o History} condition excludes history,
\textit{Always-History} incorporates recent prior visits
without relevance filtering, and ARMED incorporates only
visits retained by its history relevance gate.
All three conditions use the same ARMED pipeline,
differing only in same-patient history usage. Including history improved recommendation performance,
while relevance filtering yielded a small additional gain.

The gate removed 6 visits from 5 patients.
The 342 retained visits shared an average of 6.62 medication
labels with the current admission's ground truth, compared
with 1.00 for removed visits. Among the 5 affected patients,
mean F1 increased from 0.3632 to 0.3908, with no decrease
in F1 for any patient.
Together, these findings suggest that most available prior
visits provided relevant medication context in our test set,
while selective filtering helped exclude less relevant history.

\subsection{Retrieval Policy Ablation}

\begin{table}[t]
\caption{Retrieval policy ablation without same-patient history.}
\label{tab:retrieval_ablation}

\centering
\setlength{\tabcolsep}{4.1pt}
\begin{tabular}{lccc}
\toprule
Method & F1 & GT Coverage & Evidence Precision \\
\midrule
Single-query only & 0.4178 & 0.7174 & 0.2406 \\
Multi-query only & 0.3943 & 0.6329 & \textbf{0.2690} \\
Adaptive (ARMED) & \textbf{0.4417} & \textbf{0.7825} & 0.2437 \\
\bottomrule
\end{tabular}

\end{table}

Table~\ref{tab:retrieval_ablation} compares retrieval policies
with same-patient history disabled. All policies share the same backbone, retrieval corpus,
and no-history ARMED pipeline, differing only in retrieval
policy. Unlike the simple Static RAG baseline, which directly
provides densely retrieved cases to the LLM, these ablations
retain ARMED's reranking, structured evidence aggregation,
and label-constrained generation.
GT Coverage measures the fraction of ground-truth medication
labels covered by retrieved evidence, while Evidence Precision
measures the fraction of retrieved medication labels matching
the ground truth. Both metrics are averaged over admissions using the union
of retrieved medication labels, with empty retrievals scored
as zero. They assess retrieved evidence, not final predictions.

The \textit{single-query only} baseline retrieves using the
complete diagnosis list as one query. The \textit{multi-query
only} baseline always decomposes the diagnosis list into
focused queries and retrieves evidence for each.
The \textit{adaptive retrieval} policy proposed in ARMED
retains the initial single-query results when sufficient
and otherwise invokes multi-query retrieval to supplement
or replace them.

Always applying multi-query retrieval yielded the highest
Evidence Precision but the lowest GT Coverage and F1,
suggesting that more focused evidence alone does not ensure
better recommendations. Adaptive retrieval achieved the
highest GT Coverage and F1 while maintaining Evidence
Precision comparable to single-query retrieval. These results
support selectively expanding insufficient retrieval to
improve evidence coverage and recommendation performance.

\begin{table}[t]
\caption{Performance and runtime comparison without same-patient history.}
\label{tab:runtime_comparison}

\centering
\setlength{\tabcolsep}{7pt}
\begin{tabular}{lcc}
\toprule
Metric & Adaptive & Multi-query only \\
\midrule
F1 &
\textbf{0.4417} &
0.3943 \\

Mean latency (s) &
\textbf{14.686} &
16.865 \\

P50 latency (s) &
\textbf{12.088} &
13.466 \\

P95 latency (s) &
\textbf{28.630} &
33.934 \\

Throughput (/h) &
\textbf{245.1} &
213.5 \\
\bottomrule
\end{tabular}
\end{table}

Table~\ref{tab:runtime_comparison} additionally compares
runtime under the same settings. Relative to
\textit{multi-query only}, \textit{adaptive retrieval}
reduced mean latency by 12.9\% and increased batch-size-1
sequential throughput by 14.8\%, while achieving higher F1.
Thus, selective retrieval expansion improved both
recommendation performance and inference efficiency.

\subsection{Retrieval Strategy Breakdown}

\begin{table}[t]
\caption{Retrieval strategy breakdown without same-patient history.}
\label{tab:retrieval_breakdown}

\centering
\setlength{\tabcolsep}{5pt}
\begin{tabular}{lcccc}
\toprule
Route & $N$ & F1 & Precision & Recall \\
\midrule
Single-query sufficient
& 449
& \textbf{0.4787}
& \textbf{0.4587}
& \textbf{0.5881} \\

Supplement
& 448
& 0.4167
& 0.4215
& 0.4962 \\

Fallback
& 103
& 0.3888
& 0.3884
& 0.4798 \\
\bottomrule
\end{tabular}
\end{table}

Table~\ref{tab:retrieval_breakdown} summarizes ARMED's retrieval
routes with same-patient history disabled.
\textit{Single-query sufficient} uses the initial
retrieval alone when sufficient; \textit{Supplement}
retains the initial results and adds diagnosis-focused
multi-query evidence; and \textit{Fallback}
replaces insufficient initial results with multi-query evidence.
\textit{Single-query sufficient} achieved the highest
F1, precision, and recall, while \textit{Fallback}
was selected least often and had the lowest scores. Overall, 44.9\% of admissions required no retrieval expansion,
while 44.8\% used supplementation and only 10.3\% required
fallback, illustrating selective rather than uniform
multi-query expansion.

\subsection{Generalization Across Backbone Models}

\begin{table}[t]
\caption{Medication recommendation performance across backbone models.}
\label{tab:backbone_generalization}

\centering
\setlength{\tabcolsep}{4.1pt}
\begin{tabular}{lccc}
\toprule
Backbone
& Static RAG + History
& ARMED
& $\Delta$F1 \\
\midrule
MedGemma 1.5 4B
& 0.4074
& \textbf{0.4891}
& \textbf{+0.0817} \\

Qwen3-8B
& 0.4472
& \textbf{0.5103}
& \textbf{+0.0631} \\

Phi-3.5 Mini 3.8B
& 0.3687
& \textbf{0.4300}
& \textbf{+0.0613} \\

Gemma 3 4B
& 0.4323
& \textbf{0.4825}
& \textbf{+0.0502} \\

EXAONE 3.5 2.4B
& 0.4143
& \textbf{0.4554}
& +0.0411 \\

Ministral 3 8B
& 0.4900
& \textbf{0.4921}
& +0.0021 \\
\bottomrule
\end{tabular}
\end{table}

Table~\ref{tab:backbone_generalization} compares ARMED with
Static RAG + History across backbone models under the same
evaluation protocol, with $\Delta$F1 denoting ARMED's improvement.
Representative clinical and general-purpose models include
MedGemma 1.5 4B, Gemma 3 4B, Phi-3.5 Mini 3.8B,
EXAONE 3.5 2.4B, and Ministral 3 8B
\cite{medgemma,gemma3,phi35,exaone35,ministral3}.

ARMED improved F1 across all six backbones, with gains
ranging from 0.0021 for Ministral 3 8B to 0.0817 for
MedGemma 1.5 4B. These results suggest that its benefits
extend across backbone models, although their magnitude varies.

\begin{table*}[t]
    \caption{Case studies of static retrieval failure and adaptive recovery.}
    \label{tab:case_study}

    \centering
    \begin{tabular}{p{2.7cm}p{4.4cm}p{4.4cm}p{4.4cm}}
    \toprule
    \parbox[c]{2.6cm}{Case Type} & \parbox[c]{4.4cm}{\raggedright Ambiguous Comorbidity Case} & \parbox[c]{4.4cm}{\raggedright Weak Retrieval Case} & \parbox[c]{4.4cm}{\raggedright History Noise Control Case} \\
    \midrule

    \parbox[c]{2.7cm}{Clinical Description}
    &
    \parbox[c]{4.4cm}{\raggedright Advanced pulmonary malignancy with respiratory failure, renal disease, and cardiac ischemic comorbidity}
    &
    \parbox[c]{4.4cm}{\raggedright End-stage renal disease with metabolic derangements and complex renal/urologic history}
    &
    \parbox[c]{4.4cm}{\raggedright Neurologic/vascular presentation with diabetes and cardiovascular risk factors, but weakly relevant prior history}
    \\
    \midrule

    \parbox[c]{2.7cm}{Baseline Failure}
    &
    \parbox[c]{4.4cm}{\raggedright Broad retrieval caused many FP/FN errors}
    &
    \parbox[c]{4.4cm}{\raggedright Single-query evidence failed to recover sufficient treatments}
    &
    \parbox[c]{4.4cm}{\raggedright Recent history context increased FP burden}
    \\
    \midrule

    \parbox[c]{2.7cm}{ARMED Action}
    &
    \parbox[c]{4.4cm}{\raggedright Added multi-query supplement and retained relevant history}
    &
    \parbox[c]{4.4cm}{\raggedright Switched to multi-query fallback}
    &
    \parbox[c]{4.4cm}{\raggedright Rejected prior visits via history gate}
    \\
    \midrule

    \parbox[c]{2.7cm}{Evidence Change}
    &
    \parbox[c]{4.4cm}{\raggedright Recovered focus-specific candidates}
    &
    \parbox[c]{4.4cm}{\raggedright Retrieved diagnosis-focused evidence}
    &
    \parbox[c]{4.4cm}{\raggedright Suppressed noisy longitudinal context}
    \\
    \midrule

    \parbox[c]{2.7cm}{F1 Change}
    &
    \parbox[c]{4.4cm}{0.125 $\rightarrow$ 0.667}
    &
    \parbox[c]{4.4cm}{0.400 $\rightarrow$ 0.812}
    &
    \parbox[c]{4.4cm}{0.235 $\rightarrow$ 0.500}
    \\

    \bottomrule
    \end{tabular}
\end{table*}

\subsection{Qualitative Case Study: Static Retrieval Failure and Adaptive Recovery}

Table~\ref{tab:case_study} illustrates three examples of
adaptive recovery, comparing ARMED with Static RAG + History. ARMED supplemented retrieval and retained
relevant history in the \textit{Ambiguous Comorbidity Case},
switched to multi-query fallback in the \textit{Weak Retrieval
Case}, and excluded weakly relevant prior visits in the
\textit{History Noise Control Case}, improving case-level
F1 in all three examples.

\subsection{Additional Evaluation on MIMIC-III}

We additionally evaluated ARMED on MIMIC-III~\cite{mimiciii},
which contains real-world clinical records de-identified
by the dataset providers and was accessed through PhysioNet~\cite{physionet}
after completing the required training and credentialing
process and agreeing to the applicable data use agreement.
We used the same preprocessing procedure and NDC-to-ATC
mapping as for MIMIC-IV. All model and retrieval settings
were kept unchanged, with no dataset-specific tuning.
The retrieval corpus comprised 42,784 admissions from
33,385 patients, and the test set comprised 6,410 admissions
from 5,000 patients, with no patient or admission overlap
between the two sets. Of the test admissions, 1,410 had
prior same-patient history.

\begin{table}[t]
\caption{Medication recommendation performance on MIMIC-III.}
\label{tab:mimiciii}
\centering
\setlength{\tabcolsep}{3.8pt}
\begin{tabular}{lcccc}
\toprule
Method & F1 & 95\% CI & Precision & Recall \\
\midrule
Static RAG + History
& 0.4631 & [0.4590, 0.4671] & 0.4144 & 0.6179 \\
ARMED
& \textbf{0.4878} & \textbf{[0.4836, 0.4918]}
& \textbf{0.4601} & \textbf{0.6189} \\
\bottomrule
\end{tabular}
\end{table}

We computed 95\% confidence intervals and conducted paired
method comparisons using 10,000 patient-cluster bootstrap
resamples. Table~\ref{tab:mimiciii} shows that ARMED improved
admission-averaged F1 over Static RAG + History by 0.0247
(95\% CI: [0.0220, 0.0275]; $p<0.001$), with higher
precision and similar recall.
F1 increased from 0.4617 to 0.4853 on admissions without
prior history and from 0.4682 to 0.4967 on those with
prior history ($p<0.001$ in both groups), indicating that
the gains do not depend solely on same-patient history.

\section{Conclusion}
\label{sec:conclusion}

We proposed ARMED, an adaptive clinical RAG framework that
controls retrieval depth based on retrieval sufficiency and
selectively incorporates relevant same-patient history.
ARMED outperformed the evaluated RAG baselines,
including TreatRAG and PACE-RAG. Ablation analyses further
supported the benefits of adaptive retrieval and relevance-aware
same-patient history selection.

By combining small language models with adaptive retrieval,
ARMED supports local inference without external language
model APIs or task-specific fine-tuning. Application to a new
hospital would require local data preparation and validation.

Although ARMED improved agreement with recorded medication labels, this study did not establish the clinical appropriateness or safety of its recommendations. We did not evaluate drug-drug interactions (DDIs), validate recommendations against clinical guidelines, or conduct an expert clinical assessment. Moreover, the outputs are ATC level 3 medication classes rather than specific drugs with doses and administration routes, limiting their direct applicability to prescribing decisions. Performance on long-tail medication classes and generalization
to other institutions remain to be evaluated.

Future work could replace the current heuristic-based retrieval
controller with a trainable lightweight SLM agent capable of
learning more robust retrieval decisions from clinical context. Another important direction is to integrate medical guideline retrieval into the framework in order to refine retrieved evidence in a guideline-aware manner. Furthermore, future studies may evaluate ARMED on disease-specific EHR cohorts, such as diabetes cohorts, to assess its potential utility for prescription decision support in real hospital environments.

\section*{Acknowledgment}

This work was supported by Institute of Information \& communications Technology Planning \& Evaluation (IITP) grant funded by the Korea government (MSIT) (RS-2025-02263669, Development of server-level DRAM-stacked PIM solution for accelerating ultra-large AI models).

\bibliographystyle{IEEEtran}
\bibliography{references}

\end{document}